%% file: main.tex
\documentclass[10pt,twocolumn,letterpaper]{article}

\usepackage{cvpr}              
\usepackage[subtle,mathdisplays=normal,wordspacing=normal]{savetrees}
\usepackage{soul}
\usepackage[dvipsnames, table]{xcolor}
\definecolor{lightgray}{gray}{0.8}

\input{preamble}

\definecolor{lightgray}{gray}{0.9}
\definecolor{cvprblue}{rgb}{0.21,0.49,0.74}
\usepackage[pagebackref,breaklinks,colorlinks,citecolor=cvprblue]{hyperref}
\include{macros}
\newcommand\blfootnote[1]{%
  \begingroup
  \renewcommand\thefootnote{}\footnote{#1}%
  \addtocounter{footnote}{-1}%
  \endgroup
}
\def\paperID{6445} 
\def\confName{CVPR}
\def\confYear{2024}

\title{SDDGR: Stable Diffusion-based Deep Generative Replay \\for Class Incremental Object Detection}

\author{Junsu Kim\textsuperscript{1} \quad Hoseong Cho \textsuperscript{1,2\dag} \quad Jihyeon Kim\textsuperscript{1,3\dag} \quad Yihalem Yimolal Tiruneh\textsuperscript{1} \quad Seungryul Baek\textsuperscript{1} \vspace{0.3em} \\
{\normalsize $^1$UNIST} \qquad 
{\normalsize $^2$LG Electronics} \qquad
{\normalsize $^3$KETI}
}

\begin{document}
\maketitle
\input{sec/0_abstract}    
\blfootnote{This research was conducted when Hoseong Cho and Jihyeon Kim were graduate students (Master candidates) at UNIST$\dag$.}
\input{sec/1_intro}

\input{sec/2_RW}

\input{sec/3_Methods}

\input{sec/4_experiments}
\input{sec/5_results}

\noindent \textbf{Acknowledgements.} This work was supported by IITP grants (No. 2020-0-01336 Artificial intelligence graduate school program (UNIST) 10\%; No. 2021-0-02068 AI innovation hub 10\%; No. 2022-0-00264 Comprehensive video understanding and generation with knowledge-based deep logic neural network 10\%) and the NRF grant (No. RS-2023-00252630 10\%), all funded by the Korean government (MSIT). This work was also supported by Korea Institute of Marine Science \& Technology Promotion (KIMST) funded by Ministry of Oceans and Fisheries (RS-2022-KS221674) 20\% and received support from LG Electronics (20\%) and AI Center, CJ Corporation (20\%).
\newpage

{
    \small
    \bibliographystyle{ieeenat_fullname}
    \bibliography{main}
}

\end{document}


\maketitle

\blfootnote{This research was conducted when Hoseong Cho and Jihyeon Kim were graduate students (Master candidates) at UNIST$\dag$.}
In the supplementary material, we provide additional implementation details in Sec.~\ref{supple_sec:implementations}, extended ablation study in Sec.~\ref{supple_sec:add_ablations}, and additional experiments in Sec.~\ref{supple_sec:add_experiments}. The overall procedure is outlined in Sec.~\ref{supple_sec:procedure}. For a more comprehensive understanding of our method, refer to the additional visualizations in Sec.~\ref{supple_sec:add_visualization}. Lastly, we talk about the effect of synthetic images containing multi-object to the incremental detector and future work in Sec.~\ref{supple_sec:disscussion}

\noindent \textbf{Summary of the main paper.} In the main paper, we introduced the stable diffusion (SD)-based replay method for class incremental object detection (CIOD). Despite the SD's advanced capabilities in image generation, its direct application to CIOD presents non-trivial challenges that we address through our contributions: (1) We first enhanced the SD's grounding capabilities following GLIGEN~\cite{li2023gligen}. (2) SD often generates low-quality images and we propose an `iterative class-wise refiner' that selects $N$ quality samples for each class, by filtering out low quality images.
(3) Generated images are not always well-aligned with the bounding boxes that were inputted as the grounding inputs. If non-aligned bounding boxes are used as ground-truths in the supervised setting, it can spoil overall CIOD accuracy. We bypass the issue by involving the `L2 distillation loss': applying pre-trained network to obtain responses from generated images and enforcing the responses to a new model, without use of input bounding boxes.
~(4) We further optimized SD by adjusting text prompts and grounding signals, aiming for higher fidelity in generated images. This refinement process underscores our efforts to decrease dependency on SD's inherent capabilities.


\section{Additional implementation details}
\label{supple_sec:implementations}
\noindent\textbf{Hyper-parameters.}
In our implementation, we utilized distributed data parallel in PyTorch~\cite{paszke2019pytorch} using 4 GPUs. The total batch size is 32 in the training process. During the generation process, each GPU utilizes 4 random noises (i.e. batch size of 4 on each GPU), allowing for the simultaneous generation of 16 images. For the stable diffusion sampler, we employ the PLMS~\cite{liu2022PLMS} method. 
Additionally, we allocated both positive and negative prompts to control generated outputs.
The \emph{scene environments} of the custom positive prompt in Sec. 4.1 (main paper) used \emph{\{``demonstrating ultra high detail, 4K, 8K, ultra-realistic, crisp edges, smooth, hyper-detailed textures"\}}.
The negative prompt used \emph{\{``blurry, overlapping objects, distorted proportions, monochrome, grayscale, bad hands, deformed, lowres, error, normal quality, watermark, duplicate, worst quality, obscured faces, low visibility, unnatural colors, long body, bad anatomy, missing fingers, extra digit, fewer digits, cropped"\}}. All other hyper-parameters related to the generation carefully followed the settings of~\cite{li2023gligen}.


\begin{table}[t]
\centering
\small

\caption{Ablation study of generation-refinement on COCO 2017 (two-phase setting, 70+10). The experiment was conducted excluding L2 knowledge distillation to solely evaluate the effect of synthetic data. The best result is highlighted in \textbf{bold}. A red upward arrow \textcolor{red}{↑} indicates the performance improvement.}
\renewcommand{\arraystretch}{1.0}   
\resizebox{0.35\textwidth}{!}{%
\label{tab:sub_Infinity_versus_50}
\begin{tabular}{ccccccc}
    \toprule
        Setting & $AP$ & $AP_{.5}$ & $AP_{.75}$\\
    \midrule
        $\mathcal{N} = \infty $ & 33.4 & 50.5 & 36.4  \\
        \rowcolor{lightgray} $\mathcal{N}=50$ & \textbf{39.8} \textcolor{red}{6.4↑} & \textbf{57.7} \textcolor{red}{7.2↑} & \textbf{43.4} \textcolor{red}{7.0↑} \\
        \bottomrule
\end{tabular}}
\end{table}

\begin{table}[t]
\small
\centering
\renewcommand{\arraystretch}{1.0}
\caption{Approximate time computation based on the use of 4 A100 GPUs.}
\label{table:supple_time_computation}
\resizebox{0.46\textwidth}{!}{%
\begin{tabular}{|c|c|c|c|}
\hline
    \textbf{Class-wise counts} & \textbf{Generation time} & \textbf{Training time} & \textbf{Total time} \\
\hline
    $\mathcal{N}=\infty$ & $\sim$84 hours & $\sim$42 hours & $\sim$126 hours \\
    $\mathcal{N}=50$ & $\sim$15 hours & $\sim$21 hours & $\sim$36 hours \\
\hline
\end{tabular}}
\end{table}

\begin{algorithm*}[t]
\SetAlgoLined
\setstretch{0.85}
\small
\textbf{Input: }Generated (synthetic) dataset $\mathcal{D}_\text{gen}$, Real new dataset $\mathcal{D}_m$ for new task $\mathcal{T}_m$, Old model $\mathcal{M}_{m-1}$, New model $\mathcal{M}_m$, Old tasks $\mathcal{T}_{m-1}$. \\
\textbf{Define:} $m$: task index.  \\
\textbf{Define:} $\mathbf{y}^h$: Annotation for the $h$-th image, $\mathbf{x}^h$: $h$-th image.  \\
\textbf{Define:} $H$: The total number of data in $\mathcal{D}_\text{total}$. \\
\vspace{0.4em}

$\mathcal{D}_\text{gen} \gets \text{generation process of old tasks } \mathcal{T}_{m-1}$ \textbf{if} $m > 1$  \hfill// \textcolor{blue}{Create generated dataset. Refer to main paper Sec. 4.1 and 4.2.} \\
$\mathcal{D}_\text{total} \gets \mathcal{D}_\text{gen} \cup \mathcal{D}_{m}  \textbf{ if } m > 1$ \textbf{else} $\mathcal{D}_m$ \hfill// \textcolor{blue}{Total dataset of $\mathcal{T}_m$.} \\
\For{$h = 1, \ldots, H$}{
    Extract $h$-th annotation $\mathbf{y}_\text{total}^h$ and image $\mathbf{x}_\text{total}^h$ from $\mathcal{D}_\text{total}$. \\
    \eIf{$\mathbf{x}_\text{total}^h \in \mathcal{D}_\text{gen}$}{
        Apply L2 knowledge distillation from $\mathcal{M}_{m-1}$ to $\mathcal{M}_{m}$. \hfill // \textcolor{blue}{Training for synthetic dataset $\mathcal{D}_\text{gen}$. Refer to main paper Sec. 4.4.} \\
    }{
        Apply pseudo-labeling to $\mathbf{x}_\text{total}^h$ using $\mathcal{M}_{m-1}$. \hfill // \textcolor{blue}{pseudo-labeling to new task images ($\mathbf{x}_\text{total}^h \in \mathcal{D}_\text{m}$). Refer to main paper Sec. 4.3.} \\
        Training $\mathcal{M}_{m}$ with $\mathbf{x}_\text{total}^h$. \hfill // \textcolor{blue}{Training with pseudo labeled annotations and now annotations.} 
    }
}
\caption{Training with generated images and real data.}
\label{Alg:TrainingWithGeneratedImages}
\end{algorithm*}

\section{Additional ablations}
\label{supple_sec:add_ablations}
\noindent\textbf{Class-wise $\mathcal{N}$ limitation.}
We further evaluate the impact of the hyper-parameter $\mathcal{N}$ that limits the number of synthesized images per each class. From Tab.~\ref{tab:sub_Infinity_versus_50}, we can observe that when synthetic data is generated using the entire set of previous annotations $\mathcal{Y}_{m-1}$ ($\mathcal{N} = \infty$), the new model $\mathcal{M}_m$ tends to overfit to the synthetic data, and thereby exhibits a notable accuracy drop. Furthermore, as shown in Tab.~\ref{table:supple_time_computation}, it takes approximately $126$ hours for overall training process. In contrast, when $\mathcal{N}=50$ is used, the time is significantly reduced to $36$ hours, while achieving the best results. This implies the importance of choosing the proper number of synthetic samples during training.

\begin{table}[t]
\centering
\small
\caption{Additional comparison of different methods in $40+40$ setting on D-DETR baseline. The best result is highlighted in \textbf{bold}.  
}
\label{tab:sup_comparison_40+40}
\vspace{-1em}
\begin{tabular}{ccccccc}
    \toprule
    Method & $AP$ & $AP_{.5}$ & $AP_{.75}$ \\
    \midrule
    LWF~\cite{li2017learning} applied on D-DETR & 18.2 & 26.5 & 19.8  \\
    ICaRL~\cite{rebuffi2017icarl} applied on D-DETR & 29.9 & 42.7 & 32.7 \\
    Incremental-DETR~\cite{dong2023incrementaldetr} & 37.3 & 56.6 & - \\
    CL-DETR~\cite{liu2023CLDETR} & 42.0 & 60.1 & 45.9  \\
    \rowcolor{lightgray}SDDGR(Ours) & \textbf{43.0} & \textbf{62.1} & \textbf{47.1} \\
    \bottomrule
\end{tabular}
\end{table}
\begin{table}[t]
\centering
\small
\caption{Additional comparison of different methods in $70+10$ setting on D-DETR baseline. The ”-” symbol indicates a missing value, as reported in paper~\cite{dong2023incrementaldetr}. The best result is highlighted in \textbf{bold}.}
\label{tab:sup_comparison_70+10}
\vspace{-1em}
\begin{tabular}{ccccccc}
    \toprule
    Method & $AP$ & $AP_{.5}$ & $AP_{.75}$ \\
    \midrule
    LWF~\cite{li2017learning} applied on D-DETR & 4.3 & 6.2 & 4.8  \\
    ICaRL~\cite{rebuffi2017icarl} applied on D-DETR & 18.3 & 28.1 & 19.6 \\
    Incremental-DETR~\cite{dong2023incrementaldetr} & - & - & - \\
    CL-DETR~\cite{liu2023CLDETR} & 40.4 & 58.0 & 43.9  \\
    \rowcolor{lightgray}SDDGR(Ours) & \textbf{40.9} & \textbf{59.5} & \textbf{44.8} \\
    \bottomrule
\end{tabular}
\end{table}

\section{Additional experiements}
\label{supple_sec:add_experiments}
We provide an additional comparison of our SDDGR model with several prominent methods. These include LWF, ICARL, Incremental-DETR, and the previous state-of-the-art, CL-DETR, all of which are based on the Deformable-DETR framework. Tables ~\ref{tab:sup_comparison_40+40} and ~\ref{tab:sup_comparison_70+10} illustrate the superior performance of our SDDGR approach in both the $40+40$ and $70+10$ settings. These results highlight the effectiveness of the SDDGR method, which outperforms other methodologies in the same baseline.

\section{Overall procedure}
\label{supple_sec:procedure}
The overall procedure of the SDDGR framework is summarized in Alg.~\ref{Alg:TrainingWithGeneratedImages} to describe the sequence of synthetic image generation and training.

\section{Additional visualization}
\label{supple_sec:add_visualization}
\begin{figure*}[ht!]
\small
\centering{
\includegraphics[width=0.90\linewidth]{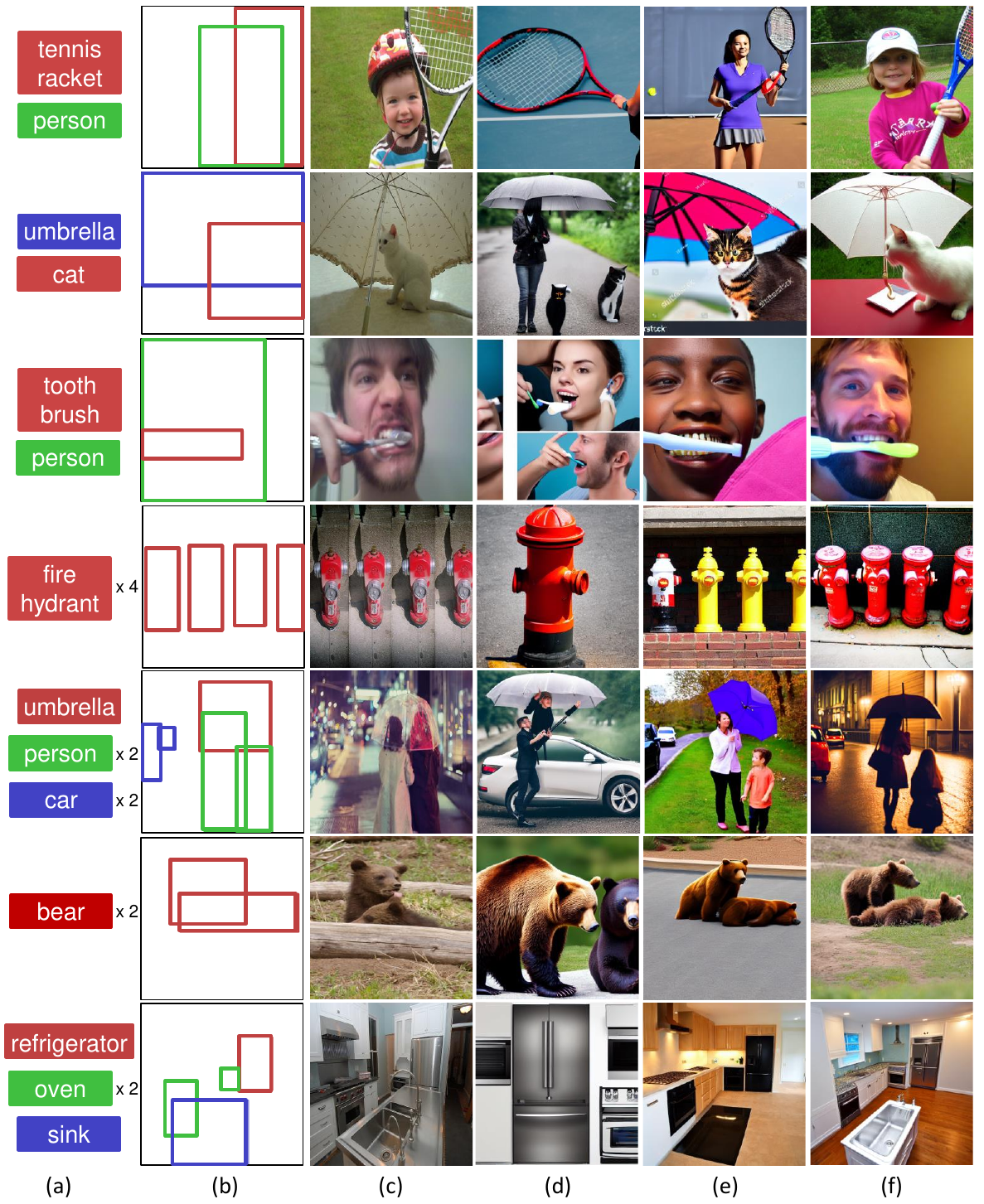}

}
    \vspace{-1em}
    \tiny\caption{Differences in image generation based on input types. Each row in this figure represents different examples used for image synthesis, with varying prompts. The first row uses prompts `A photo of tennis racket and person, realistic, ... details.' The second row uses prompts `A photo of umbrella and cat, ...'. The third row uses prompts `A photo of toothbrush and person, ...'. The fourth row uses prompts `A photo of four fire hydrants, ...'. The fifth row uses prompts `A photo of umbrella, two persons, and two cars, ...'. The sixth row uses prompts `A photo of two bears, ...'. Finally, the last row uses prompts `A photo of a refrigerator, two ovens, and a sink, ...'. For each set of images: (a) and (b) show the grounding inputs (object classes and bounding boxes), (c) displays COCO real images, (d) depicts images generated by inputting only the text prompt, (e) shows images generated by combining grounding inputs and the text prompt, and (f) illustrates images generated by incorporating the combination of text prompt, grounding inputs, and CLIP's image embedding.}
\label{fig:supple_various_inputs}
\end{figure*}

\begin{figure*}[t!]
\small
\centering{
\includegraphics[width=\linewidth]{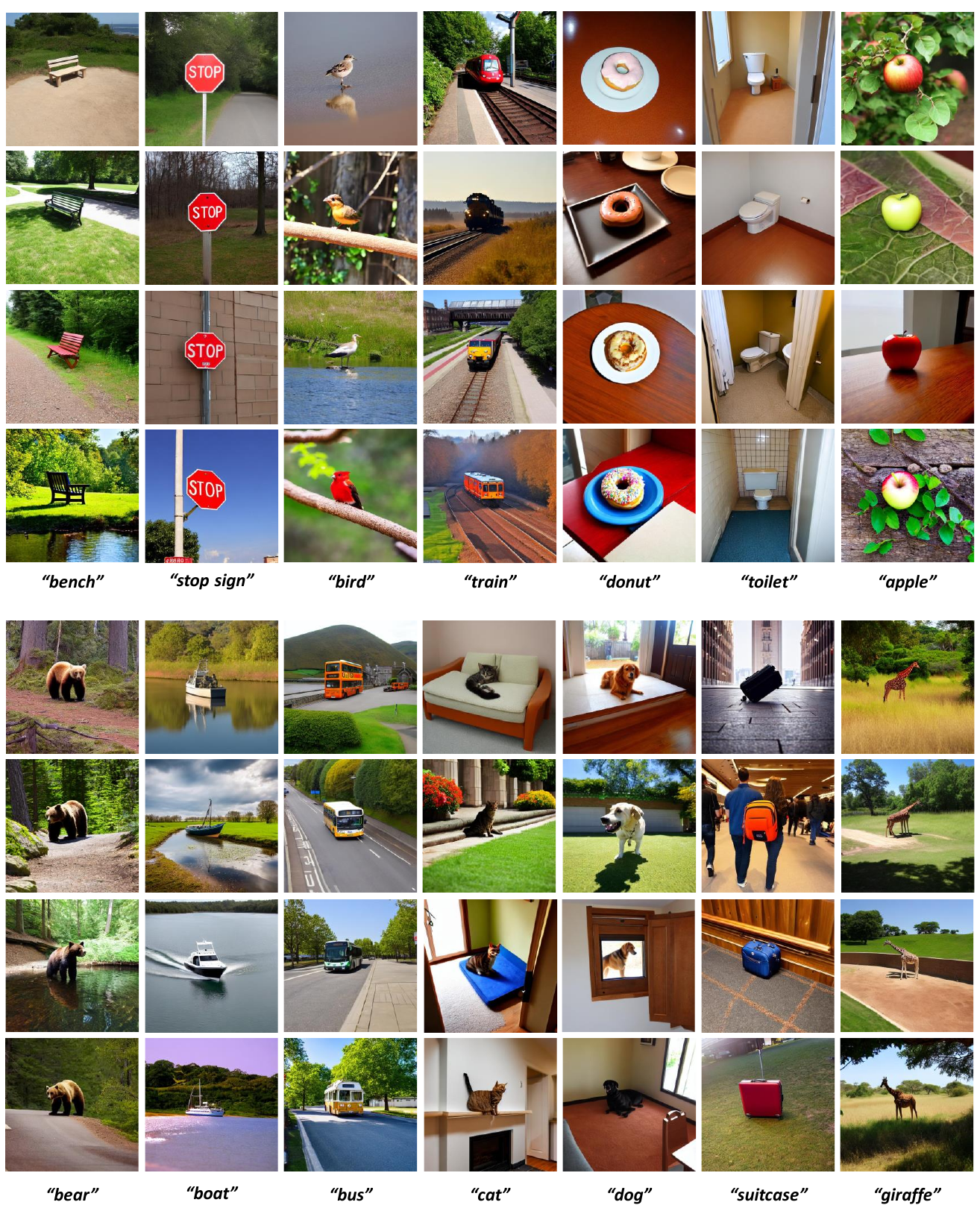}

}
    \vspace{-1em}
    \tiny\caption{Example generated images for the class-specific generation process.}
\label{fig:supple_mono_objs}
\end{figure*}

\begin{figure*}[ht!]
\small
\centering{
\includegraphics[width=1\linewidth]{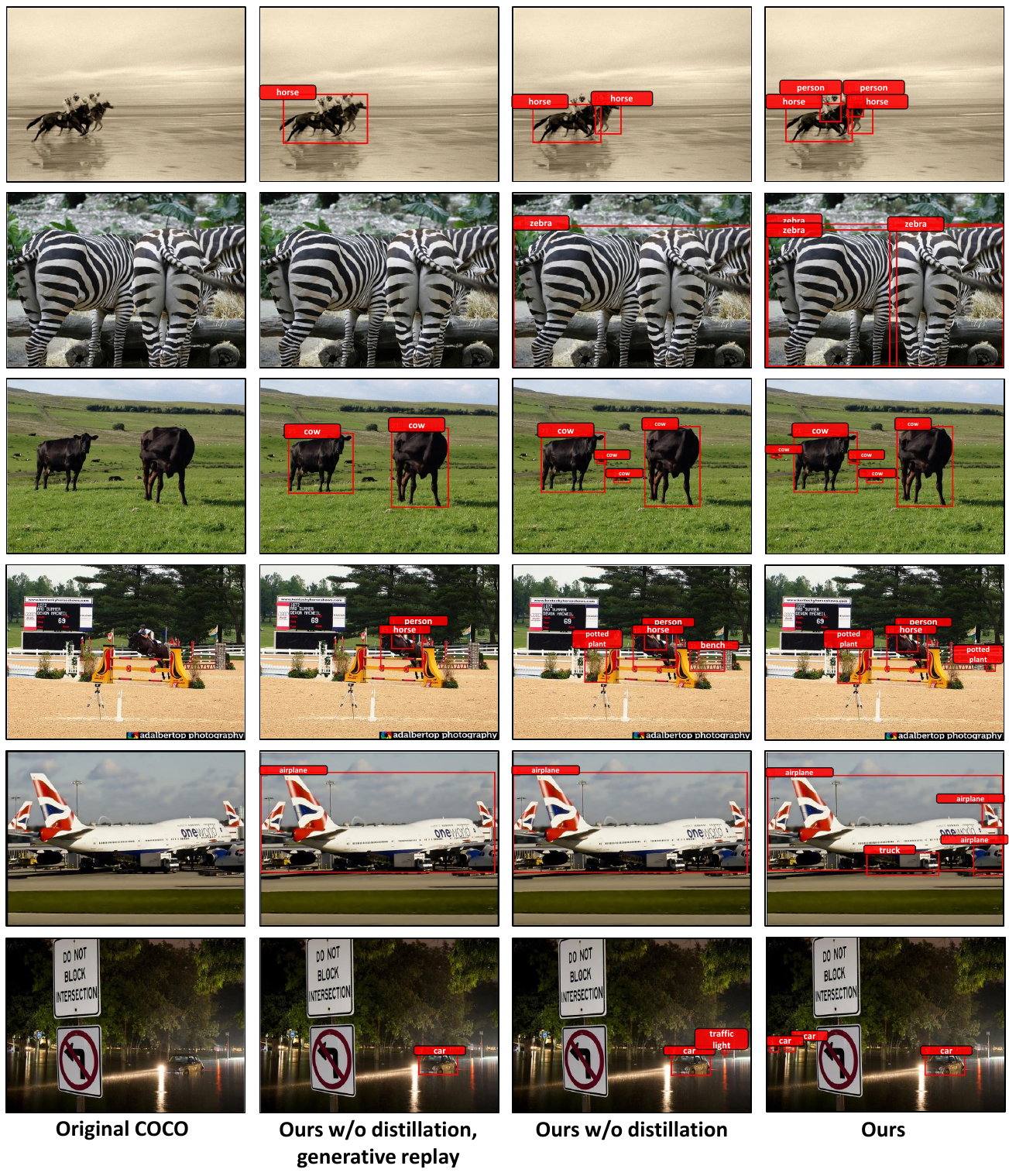}
}
    \vspace{-1em}
    \tiny\caption{Our detection results on COCO 2017 (two-phase setting, 70+10).}
\label{fig:supple_results}
\end{figure*}

\begin{figure*}[ht!]
\small
\centering{
\includegraphics[width=1\linewidth]{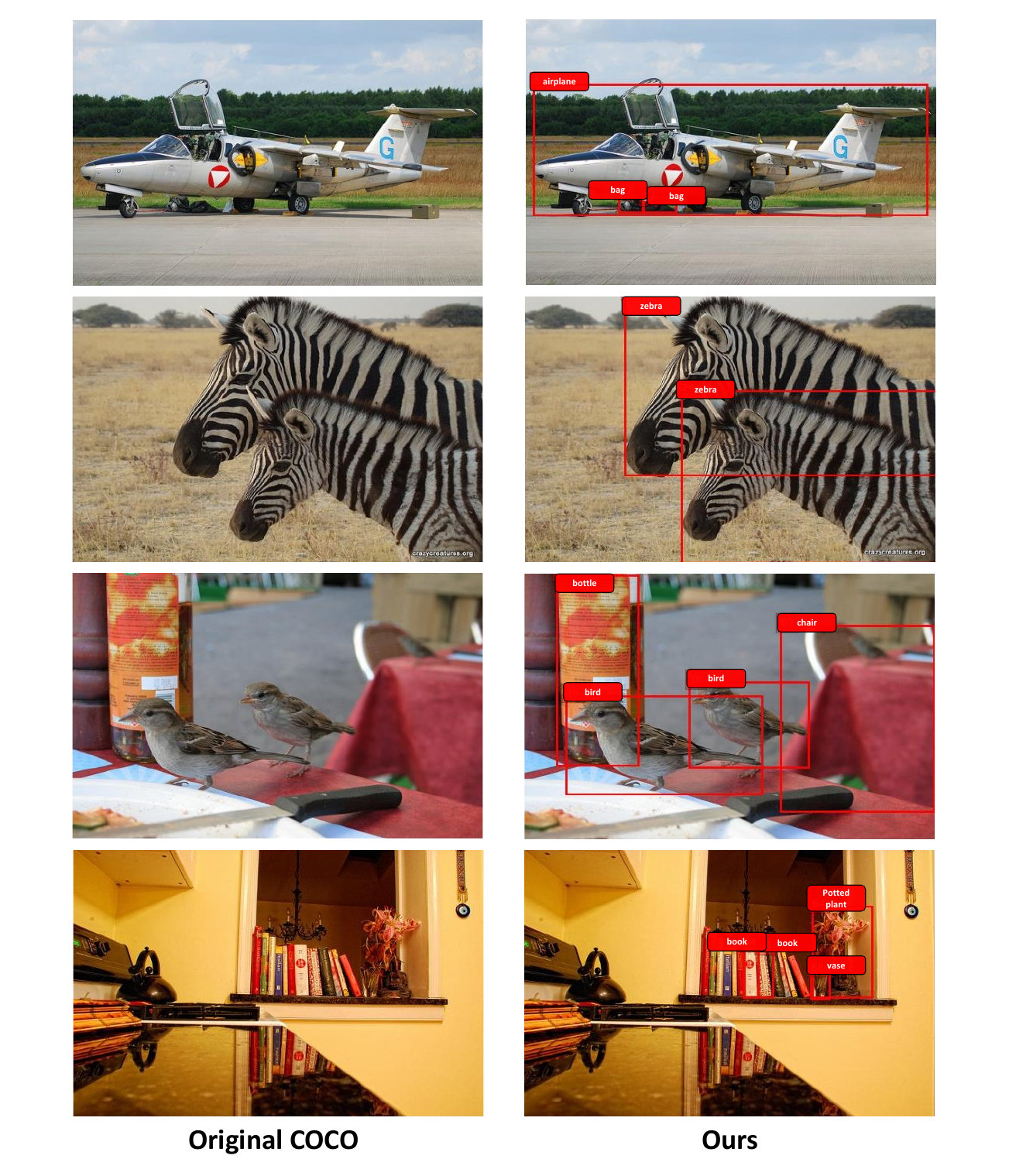}
}
    \vspace{-1em}
    \tiny\caption{Our detection results on COCO 2017 (two-phase setting, 70+10).}
\label{fig:supple_results2}
\end{figure*}

We experimented with the effect of the different inputs on the final image quality in Fig.~\ref{fig:supple_various_inputs}. We also visualized more examples for generated images according to their classes in Fig.~\ref{fig:supple_mono_objs}. Additionally, Fig.~\ref{fig:supple_results} presents the detection results using three of our baselines and Fig.~\ref{fig:supple_results2} presents the more detection results of ours.
\begin{figure}[t]
\small
\vspace{-1.8em}
\centering{
\includegraphics[width=1\linewidth]{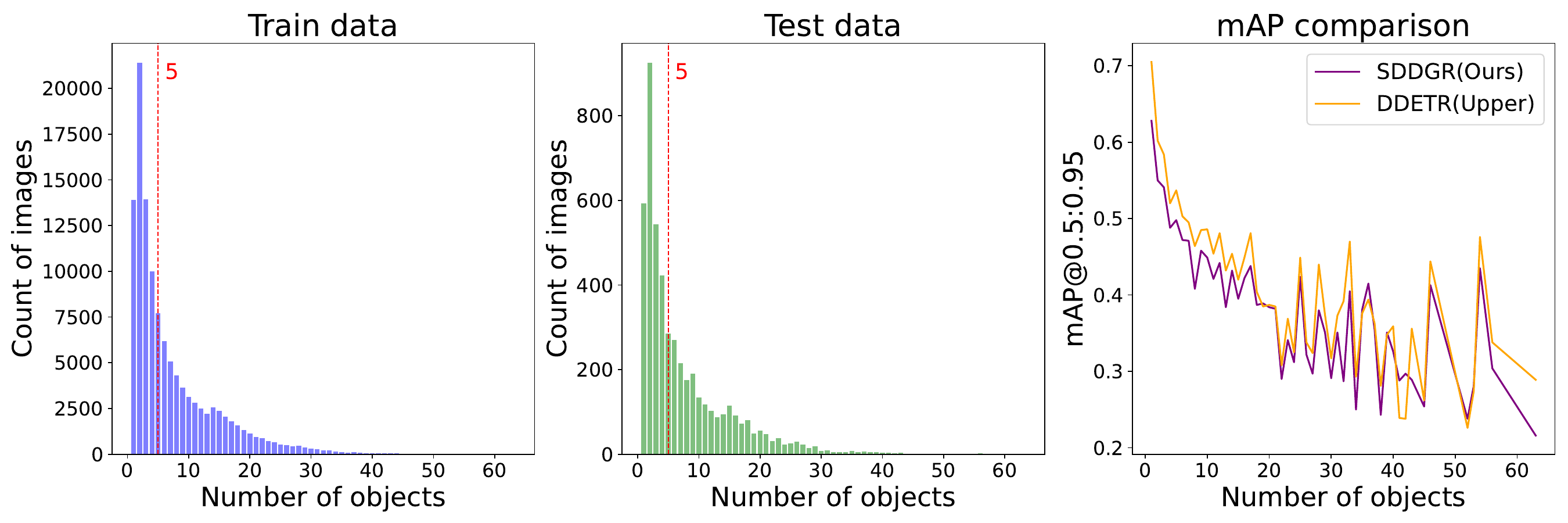}
}
    \vspace{-1.5em}
    \tiny\caption{(Col. 1-2) dist. of object \#  in train/test sets, (Col. 3) mAP gap between SDDGR(Ours) vs. DDETR according to \# of objects.}
\label{fig:reb_distribution}
\vspace{-2.1em}
\end{figure}

\section{Discussion and future work}
\label{supple_sec:disscussion}
This section explores the challenges of generating synthetic images, especially those depicting scenes with multi-objects. It is commonly believed that the performance of generative models, such as Stable Diffusion (SD), declines as the complexity of the scene increases. To directly analyze the challenges' impact on SDDGR, we have examined the distribution of images by the number of objects alongside the model's performance, and we present this analysis in Fig.~\ref{fig:reb_distribution}. 

Essentially, there is a sharp decrease in the frequency of images containing more than five objects (denoted by the \textcolor{red}{red} line) on COCO dataset distribution, as depicted on the leftmost and middle of Fig.~\ref{fig:reb_distribution}. Furthermore, the rightmost of Fig.~\ref{fig:reb_distribution}, contrasts the mean Average Precision (mAP) of our SDDGR model in a 70+10 setting with that of the upper-bounded Deformable-DETR (D-DETR), which trained with integrated both new and older classes. From the graph, we found that the accuracy disparity between SDDGR and the upper-bounded D-DETR is minimal across different object counts. This observation suggests that the inherent challenges of complex scene generation do not significantly affect SDDGR's performance. However, we think explicitly tackling the aspect could be a future research direction.



{
    \small
    \bibliographystyle{ieeenat_fullname}
    \bibliography{main}
}

%% file: preamble.tex
\usepackage[normalem]{ulem}
\usepackage{multirow}
\usepackage{graphicx}
\usepackage{subcaption}
\usepackage{amsmath}

\usepackage{colortbl}
\usepackage{booktabs}
\usepackage{amssymb} 
\usepackage{tikz}
\usepackage[linesnumbered,ruled,vlined]{algorithm2e}
\usepackage{algpseudocode}
\SetKwInput{KwInput}{Input}                
\SetKwInput{KwDefine}{Define}                
\SetKwInput{KwOutput}{Output}              
\SetKwInput{KwInitialize}{Initialize}              

\usepackage{setspace}                      
\usepackage[accsupp]{axessibility}
\usepackage{pifont} 


%% file: macros.tex
\usepackage{mathtools}
\usepackage{xspace}

\newcommand{\ignore}[1]{}

\makeatletter
\DeclareRobustCommand\onedot{\futurelet\@let@token\@onedot}
\def\@onedot{\ifx\@let@token.\else.\null\fi\xspace}
\makeatother

%% file: sec/0_abstract.tex
\begin{abstract} 
In the field of class incremental learning (CIL), generative replay has become increasingly prominent as a method to mitigate the catastrophic forgetting, alongside the continuous improvements in generative models. However, its application in class incremental object detection (CIOD) has been significantly limited, primarily due to the complexities of scenes involving multiple labels. In this paper, we propose a novel approach called stable diffusion deep generative replay (SDDGR) for CIOD. Our method utilizes a diffusion-based generative model with pre-trained text-to-image diffusion networks to generate realistic and diverse synthetic images. SDDGR incorporates an iterative refinement strategy to produce high-quality images encompassing old classes. Additionally, we adopt an L2 knowledge distillation technique to improve the retention of prior knowledge in synthetic images. Furthermore, our approach includes pseudo-labeling for old objects within new task images, preventing misclassification as background elements. Extensive experiments on the COCO 2017 dataset demonstrate that SDDGR significantly outperforms existing algorithms, achieving a new state-of-the-art in various CIOD scenarios. 
\end{abstract}

%% file: sec/1_intro.tex
\section{Introduction}
\label{sec:intro}
The key challenge in artificial intelligence is the development of models capable of continuous learning, similar to human knowledge accumulation over a lifetime. This challenge has sparked the field of class incremental learning (CIL), the continual learning in the classification task. The CIL focuses on developing techniques that enable models to learn new classes without compromising previously acquired knowledge. 

To address the challenge, researchers have primarily focused on mainly two strategies: knowledge distillation~\cite{li2017learning, hao2019end, simon2021learning, lu2022augmented, rebuffi2017icarl}, replay~\cite{rebuffi2017icarl, shin2017continual, he2018exemplar, chaudhry2019continual, de2021continual, wu2018MRGAN, cong2020ganmemory, xiang2019ILCAN, shin2017dgr, gao2023ddgr}. Among these, replay has been employed as a prominent solution in addressing the challenge of forgetting. Replay can be classified into two categories: partial experience replay~\cite{rebuffi2017icarl, shin2017continual, he2018exemplar, chaudhry2019continual, de2021continual} and deep generative replay~\cite{wu2018MRGAN, cong2020ganmemory, xiang2019ILCAN, shin2017dgr, gao2023ddgr}. The partial experience replay needs to store actual data samples from old tasks, acting as a reservoir of previous knowledge for the model. On the other hand, generative replay employs generative models to mimic the distribution of old task's data, enabling the current model to re-experience the previous knowledge.

\begin{figure}[t]
\centering{
\includegraphics[width=1.\linewidth]{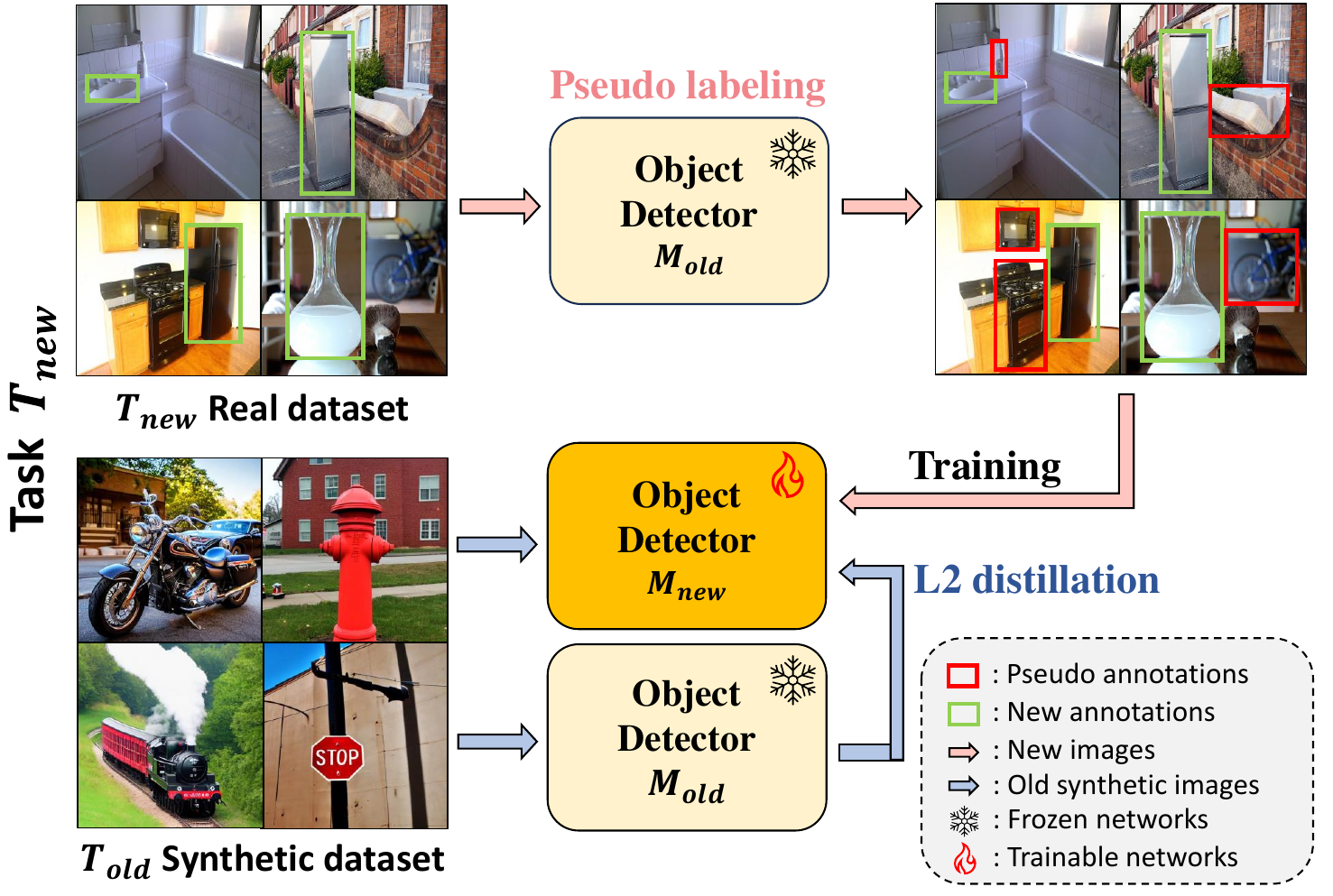}
}
    \vspace{-1em}
    \caption{We utilize a pre-trained text-to-image diffusion model~\cite{rombach2022high} to generate realistic images that include objects from the old task. These images are then filtered out via iterative refinement and filtered synthetic images are integrated into the training process of the new task. During training, we employ L2 distillation to a synthetic dataset. Additionally, when training an image for the new task, we employ a pseudo-labeling that finds the old task objects from the new task images. The series of methods enable us to effectively mitigate the issue of catastrophic forgetting.}
\label{fig:fig1}
\vspace{-1em}
\end{figure}
These methods have made significant progress in the field of image classification when there is only a single object present in an image. Yet, there has been a pressing need for techniques that can handle more complex and realistic scenes including multi-labels in a scene based on the object detection algorithms. Consequently, class incremental object detection (CIOD) has emerged, with the goal of improving models to detect multiple labels in a scene while still preserving the ability to recognize previously learned object classes.

Initial researches~\cite{shmelkov2017incremental,liu2020incdet, acharya2020rodeo, feng2022erders} in class incremental object detection (CIOD) extended image classification methods to CIOD, showing encouraging results. 
Furthermore, as the Transformer-based architectures~\cite{carion2020end, zhu2020deformable, li2022dn} are introduced as the alternative to the CNN-based approaches~\cite{li2020generalized, ren2015faster}, the CIOD for Transformer-based object detector is also proposed. Specifically, Gupta~\etal~\cite{gupta2022ow} and Liu~\etal~\cite{liu2023CLDETR}, which utilize deformable-DETR~\cite{zhu2020deformable}, have introduced distinctive characteristics into Transformer-based object detection, also incorporating partial experience replay in their methodologies. 
Despite significant advancements, they still heavily rely on the direct use of real data. 


In parallel to the advancements in incremental learning, generative models have seen noticeable advancements. Moving away from traditional generative models like generative adversarial networks (GANs)~\cite{goodfellow2014generative,mirza2014conditionalgan} and Variational autoencoders (VAEs)~\cite{kingma2013auto}, recent image generation has focused on more sophisticated and realistic techniques, such as diffusion models~\cite{ho2020denoising, dhariwal2021diffusion, song2020denoising, song2020score}. 
Notably, the stable diffusion (SD) method~\cite{rombach2022high}, which has been trained on a vast amount of online knowledge~\cite{schuhmann2021laion, schuhmann2022laion}, has gained significant attention for its impressive performance. This has led to various studies~\cite{li2023gligen, zhang2023controlnet, voynov2023sketch, yang2023reco} to utilize the model's capabilities with its original weights fixed. Motivated by the research trends, we proposed to utilize the pre-trained SD network for high-quality image generation to prevent the catastrophic forgetting. While generative models like the SD have shown proficiency in reproducing knowledge from the prompt, their effectiveness in multi-label scenarios, such as CIOD, remains constrained by the complexity of the scenario. However, we observed that the na\"ive application of SD is not suitable for successful CIOD. Thus, we proposed to improve the SD to control it based on grounding inputs such as classes and bounding boxes, via GLIGEN~\cite{li2023gligen}. Furthermore, we proposed series of methods to secure the generated image quality.



In this study, we introduce the stable diffusion-based deep generative replay (SDDGR) strategy, a novel method for utilizing a pre-trained generative model for mitigating the catastrophic forgetting in CIOD. The SDDGR generates images by using grounding inputs and prompts that explain complex scenes, which include previously learned objects. 
However, we observed that the pre-trained SD weights are sub-optimal for the CIOD. To relieve the issue, we further proposed to refine the image fidelity through iterative refinement via a trained detector. Additionally, we trained a model using the L2 distillation to facilitate effective knowledge transfer from these synthetic images to the updated model, rather than the direct training. Simultaneously, we perform the pseudo-labeling for the old task's objects which exist in the new task's training images, to prevent it from being detected as the background. Using series of proposed methods, the SDDGR demonstrates excellent performance on the COCO dataset, achieving state-of-the-art accuracy. The overall training process of our method is shown in Figure~\ref{fig:fig1}. Our contributions are summarized as follows:
\begin{itemize}
    \item As far as we are aware, we, for the first time, proposed to apply the diffusion-based generative model in CIOD problem. 
    \item Na\"ively applying the diffusion model to CIOD can decrease the overall accuracy. To make it properly work, we introduced series of methods to improve the generated image quality, to prevent the overfitting or mis-led information during training.    
    \item The extended experiments demonstrate state-of-the-art performance on the COCO dataset, substantiating its efficacy in various CIOD scenarios.
\end{itemize}

%% file: sec/2_RW.tex
\section{Related works}
\label{sec:rw}

\subsection{Continual learning}
\label{subsec:CL}
Class-incremental learning (CIL) is a subset of continual learning, aiming to seamlessly integrate new classes into a model while maintaining the ability to recognize existing ones. Most influential CIL studies focused on classification, where one image represents a single class. In our paper, unless otherwise specified, CIL refers to the classification task.
On the other hand, class incremental object detection (CIOD) presents a challenging task due to the presence of multiple instances belonging to various classes within images. When instances are trained under different tasks, they cannot be trained simultaneously. This can cause the model to classify these instances as background, which in turn degrades detection performance sequentially. Despite the clear challenges, CIOD has received relatively less research attention compared to CIL due to its complex nature.

\begin{figure*}[ht!]
\small
\centering{
\includegraphics[width=1\linewidth]{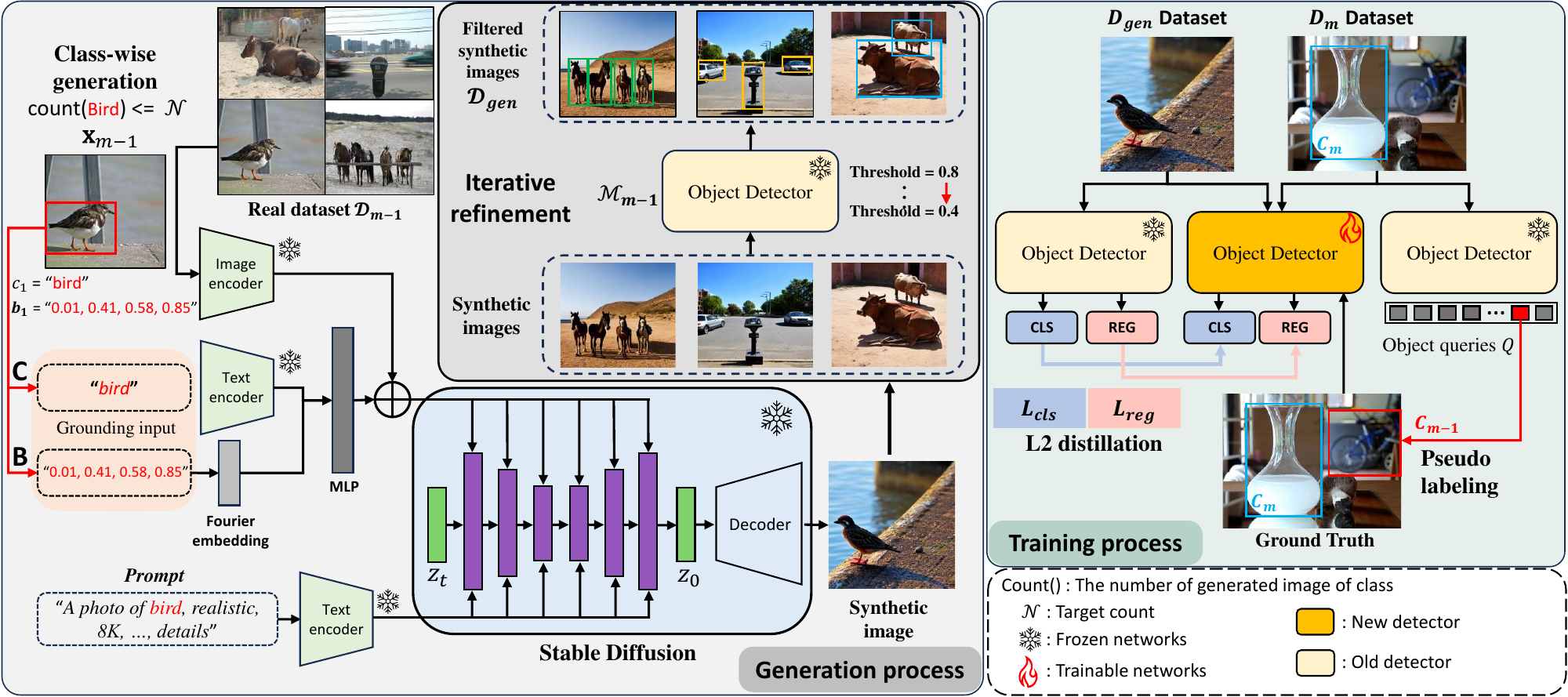}
}
    \vspace{-1em}
    \caption{\emph{Schematic of Our SDDGR Framework:} In the `Generation process', our method individually generates each image based on class labels $\mathbf{C}_\text{label}$, specific bounding box locations $\mathbf{B}_\text{location}$, and old real images $\mathbf{x}_{m-1}$ in the old dataset $\mathcal{D}_{m-1}$. An `Iterative refinement', employing the trained model $\mathcal{M}_{m-1}$, is applied to these synthetic images. In this algorithm, images with object scores below a dynamically adjusted threshold (ranging from 0.8 to 0.4 in our study) are systematically excluded. This cycle of generation and dynamic refinement continues until each class reaches the pre-defined target number of instances $\mathcal{N}$, or the lower threshold limit is met. In the `Training process', the synthetic dataset is utilized for the continual learning via L2 distillation loss. Furthermore, real images undergo pseudo-labeling before being incorporated into the `Training process'.}
    
    
\label{fig:Overall}
\end{figure*}
\vspace{0.3em}
\noindent{\textbf{Class incremental learning.}} 
In CIL, we can cluster the main methods into knowledge distillation~\cite{li2017learning, hao2019end, simon2021learning, lu2022augmented}, and replay~\cite{rebuffi2017icarl, shin2017continual, isele2018selective,  he2018exemplar, rolnick2019experience, chaudhry2019continual, de2021continual, wu2018MRGAN, cong2020ganmemory, xiang2019ILCAN, shin2017dgr, gao2023ddgr} in general.
Among these, Replay methods are most frequently utilized for their simple yet powerful effects and can be categorized into two types: partial experience replay (ER)~\cite{rebuffi2017icarl, guo2020randomsampling, bang2021rainbow, prabhu2020gdumb, koh2021blurry, chaudhry2018randomsampling} and generative replay (GR) ~\cite{gao2023ddgr, shin2017dgr, wu2018MRGAN, cong2020ganmemory, xiang2019ILCAN, jodelet2023SDCIL}. The former involves reusing a subset of the original data repeatedly, while The latter employs a generative model to recreate the data distribution of previous tasks, effectively mitigating the forgetting~\cite{robins1995catastrophic}. In the GR, DGR~\cite{shin2017dgr} is an initial attempt of the GR method to prevent the loss of incremental classes using GAN~\cite{goodfellow2014generative}. MRGAN~\cite{wu2018MRGAN} and ILCAN~\cite{xiang2019ILCAN}, which evolved from DGR. Furthermore, DDGR~\cite{gao2023ddgr} leverages advanced generative models, particularly diffusion-based techniques, to enhance the fidelity and variability of generated data. However, these methods have mainly been used in simpler scenarios for CIL because they require significant resources for training a generative. In our research, we shift the focus to applying advanced generative models within CIOD.

\vspace{0.3em}
\noindent{\textbf{Class incremental object detection.}} CIOD has progressed from primarily employing CNN-based methods~\cite{shmelkov2017incremental, feng2022erders, acharya2020rodeo, liu2020multi, peng2021sid, li2019rilod} to also incorporating Transformer-based approaches~\cite{gupta2022ow, liu2023CLDETR, dong2023incrementaldetr, kim2024vlm, kim2024class-wise}. In this trend, ILOD~\cite{shmelkov2017incremental}, a pioneering work in CIOD, implemented the LWF~\cite{li2017learning} method to handle forgetting. Besides, Feng~\etal~\cite{feng2022erders} focuses on maximizing the utility of heads in the distillation. More recent developments like CL-DETR~\cite{liu2023CLDETR} and OW-DETR~\cite{gupta2022ow} have adopted the Deformable DETR~\cite{zhu2020deformable} (D-DETR) as a baseline. CL-DETR employs knowledge distillation at the level of the labels, utilizing an old model to perform this process. OW-DETR introduce attention-driven pseudo-labeling, helping to identify unrecognized labels. In this study, we use D-DETR as a base detector to exploit the advantages of DETR~\cite{carion2020end} and compare its performance.

\subsection{Diffusion models}
Diffusion models have been largely researched due to their powerful generation capability.~\cite{ho2020denoising, song2020denoising, song2020score} proposed a basic framework for training through U-Net~\cite{ronneberger2015u}.~\cite{dhariwal2021diffusion, ho2022classifier} have demonstrated superior results compared to GAN~\cite{goodfellow2014generative} and VAE~\cite{kingma2013auto} based methods. However, since these models typically operate directly in pixel space, substantial computational resources are required. To solve this problem, Rombach~\etal~\cite{rombach2022high} proposed latent diffusion model(LDM), which performs the diffusion steps in latent space. They leveraged large-scale datasets such as LAION~\cite{schuhmann2021laion} and the pre-trained BERT~\cite{devlin2018bert} for text-to-image synthesis. This approach enables the incorporation of conditions during the image generation process, leading to the generation of desired images. Building on this foundation, Stability AI advanced the field further by developing stable diffusion (SD). SD utilizes an even larger dataset~\cite{schuhmann2022laion} and incorporates pre-trained CLIP~\cite{radford2021learning}. Recent research has focused on using pre-trained SD as a foundation to effectively leverage the extensive knowledge.~\cite{li2023gligen, zhang2023controlnet, voynov2023sketch, huang2023composer, mou2023t2i} have gained popularity for controlled generation by incorporating additional conditions.~\cite{shipard2023diversity, yang2023boosting} have demonstrated performance enhancements by generating additional data for training. In line with these advancements, our study employs pre-trained SD as a form of generative replay model to prevent the forgetting of previous knowledge.


%% file: sec/3_Methods.tex
\section{Preliminaries}
\subsection{Stable diffusion}
Stable diffusion (SD)~\cite{rombach2022high} includes an VAE~\cite{kingma2013auto} structure for first extracting the latent vector $\mathbf{z}\in\mathbb{R}^{64\times 64}$ from the image $\mathbf{x}\in\mathbb{R}^{512\times 512}$ and gaining the same dimensional reconstructed images $\hat{\mathbf{x}}$ from the latent vector $\mathbf{z}$. It uses also a U-Net~\cite{ronneberger2015u} architecture to add Gaussian noise to the latent vector and to remove the noise during the backward process, which is called the diffusion process~\cite{ho2020denoising, song2020denoising, song2020score}. By leveraging the text embedding of CLIP~\cite{radford2021clip} and cross-attention~\cite{vaswani2017attention}, SD efficiently generates images based on the text prompt $\mathbf{T}$. 

The core function is $f_{\theta}(\mathbf{z}_t, t, \mathbf{T})$, where the trained U-Net is used for $f_{\theta}$, $t$ denotes the time embedding and $\mathbf{z}_{t}$ represents the latent representation at the $t$-th diffusion time step. 
Although SD is adept at generating images from assigned prompts $\mathbf{T}$, it lacks the capability to utilize additional grounding inputs that would guide the generation process in terms of specific locations and categories of objects, thus limiting the elaboration of images.

\subsection{Controllable image generation}
To exploit the SD in the context of CIOD, we need to involve additional conditions such as bounding boxes and classes particularly when generating images with scenes containing multiple objects. However, as pointed out before, the SD lacks such a capability. To address this limitation, we extended the SD to incorporate the additional guiding inputs, following GLIGEN~\cite{li2023gligen} (Unless otherwise noted, subsequent references to SD in this paper denote the SD whose grounding capability is enhanced by the use of GLIGEN~\cite{li2023gligen}.) This approach is able to leverage the pre-trained knowledge in the SD; while using the grounding inputs, such as classes and bounding boxes (bbox) additionally to the original text prompt $\mathbf{T}$. Grounding inputs are represented as classes and bounding boxes for $N$ objects in an image as follows:
\begin{eqnarray}
\small
    \mathbf{C}_\text{label} &= [c_1, \ldots, c_N], \\
    \mathbf{B}_\text{location} &= [\mathbf{b}_1, \ldots, \mathbf{b}_N],
\end{eqnarray}
where each $c_i$ represents a specific class within the set of trained classes $C$, and $\mathbf{b}_i$ denote the corresponding bounding box's normalized coordinate values $[x_{min}, y_{min}, x_{max}, y_{max}]$ for the $i$-th instance, respectively. Now, the SD becomes to combine the text prompt $\mathbf{T}$ with grounding tokens $\mathbf{C}_\text{label}$ and $\mathbf{B}_\text{location}$ using a gated self-attention mechanism, to generate accurate images. The diffusion function $f_{\theta}$ is then modified to incorporate grounding inputs:
\begin{equation}
\small
    f_{\theta}(\mathbf{z}_t, t, \mathbf{T}, \mathbf{C}_\text{label}, \mathbf{B}_\text{location}).
\end{equation}
Furthermore, a hyper-parameter $\beta\in[0,1]$ is used to handle the influence of grounding inputs over the diffusion process. 

\section{Methods}
\label{sec:method}
The objective of class incremental object detection (CIOD) is to progressively assimilate new classes without compromising the knowledge of previously learned classes. This paradigm is characterized by a sequence of tasks, each represented as $\mathcal{T}_m$, where $m\in[1,M]$ and $M$ denote the cumulative number of tasks. Each task contains specific data, represented as $\mathcal{D}_m$. Specifically, the dataset $\mathcal{D}_m$ consists of a set of input images $\mathcal{X}_m=\{\mathbf{x}_{m}^1, \ldots, \mathbf{x}_{m}^D\}$ and a set of corresponding annotations $\mathcal{Y}_m=\{\mathbf{y}_{m}^1, \ldots, \mathbf{y}_{m}^D\}$, where $D$ is the data length. It is important to note that in object detection, individual annotations $\mathbf{y}_{m}^i$ consist of multiple object instances. We also follow the conventional CIOD configuration~\cite{shmelkov2017incremental, feng2022erders}, which implies that some images can be shared across different tasks. 

Our approach, called SDDGR, consists of four key modules: 1) A method to generate images that include previous class objects (Section~\ref{subsec:generate process}), 2) A technique for filtering more expressive images (Section~\ref{subsec:refiner}), 3) A method for implementing pseudo labeling of the DETR framework (Section~\ref{subsec:fake_query}), and 4) A training protocol for using the synthetic images (Section~\ref{subsec:training_process}). Figure~\ref{fig:Overall} provides a comprehensive overview of these components.
\subsection{Image generation}
\label{subsec:generate process}
\noindent\textbf{Text prompt Design.} 
To generate images that accurately reflect the object categories of previous tasks, we carefully design text prompts $\mathbf{T}$ that encapsulate object classes of $\mathcal{Y}_{m-1}$ from the previous dataset $\mathcal{D}_{m-1}$. Initially, we identify the object labels, as multiple objects may appear in a single image for CIOD. Subsequently, we count the occurrence of each object and express the number using words (e.g., one, two, etc.). As a result, we frame our prompts to reflect both the object category and the number of occurrences: ``A photo of \{count\} \{object A\}, \{count\} \{object B\}, and \{count\} \{object C\}, \{scene environments\}". The term $\{\text{scene environments}\}$ is included at the end of the prompt to describe the overall style and aesthetic quality of the image (e.g. 4K, 8K, realistic, etc). However, when generating images in SD with prompts, not all prompts are accurately reflected, so it is still challenging to precisely place the objects.

\noindent\textbf{Control strategy for stable diffusion.} To generate images $\mathcal{X}_\text{gen}$ that are consistent with both spatial and object categories from the previous dataset $\mathcal{D}_{m-1}$, we employ grounding input in conjunction with the text prompt $\mathbf{T}$. Specifically, for each annotation $\mathbf{y}_{m-1}^i$, we extract the category labels $\mathbf{C}_\text{label}$ and their corresponding object locations $\mathbf{B}_\text{location}$ for all entities. 
The grounding input is defined as a set of label and location pairs as follows:
\begin{align}
\small
    \{\underbrace{\mathbf{C}_{\text{label}},\mathbf{B}_{\text{location}}}_{\textit{grounding input}} \}= \{ \underbrace{(c_1, \mathbf{b}_1), (c_2, \mathbf{b}_2), \ldots, (c_N, \mathbf{b}_N)}_{\textit{entities}} \}. 
\end{align}

\begin{figure}[t]
\centering{
\includegraphics[width=1.\linewidth]{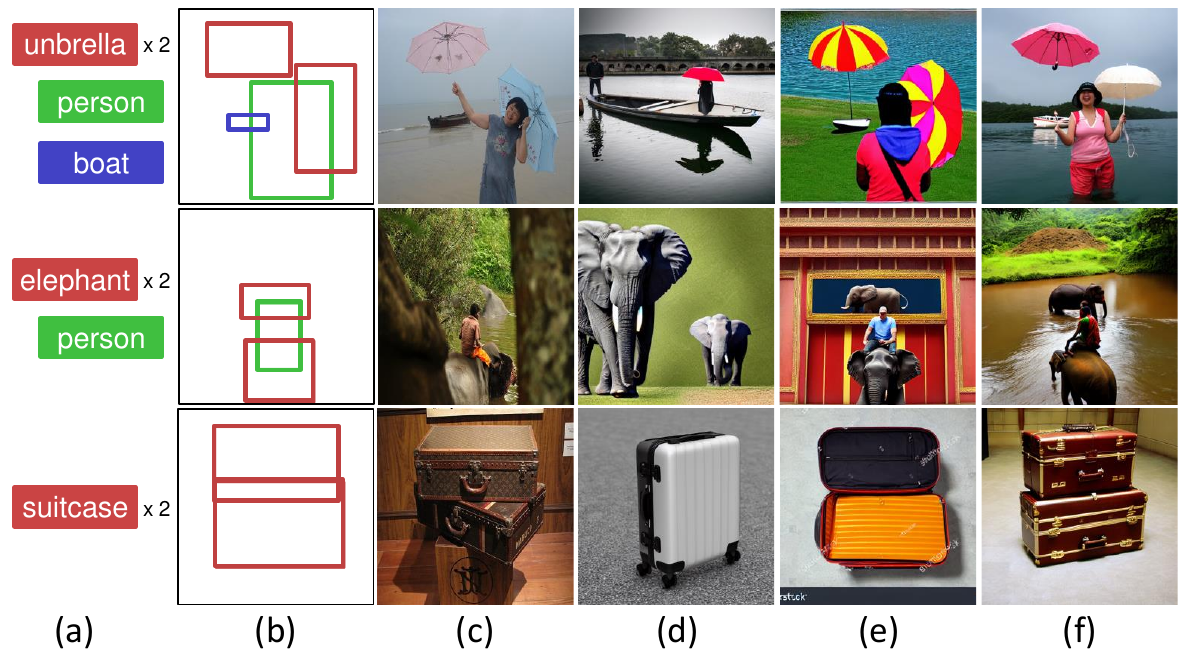}
}
    \vspace{-1em}
    \tiny\caption{Differences in image generation based on input types. Each row represents different examples used for image synthesis. The first row uses prompts like ``A photo of two umbrellas, person and boat, realistic, ... details". The second row uses prompts like ``A photo of two elephants and person, ...". The last row uses prompts like ``A photo of two suitcases, ...". (a) and (b) show the grounding input. (c) shows COCO real images. (d) depicts the prompt-only synthetic images. (e) depicts combined the grounding input and the prompt. (f) shows used the prompt, grounding input, and CLIP image embedding for image synthesis.}
\label{fig:fig3_groundinginput}
\end{figure}

  Next, we use the text encoder in CLIP~\cite{radford2021clip} to convert the labels $\mathbf{C}_\text{label}$ into text-to-image matching embeddings, the same as we apply to the prompts for the SD. Concurrently, the location $\mathbf{B}_{\text{location}}$ is transformed into the Fourier embeddings as suggested by Mildenhall~\etal~\cite{mildenhall2021nerf} for high-dimension representation. These embeddings are then fused across the feature dimension by the MLP layer, serving as a condition for the SD. The fused grounding embeddings are incorporated into the generation process using a gated self-attention fusion strategy~\cite{li2023gligen} with the text prompt $\textbf{T}$. We employ them throughout the entire denoising process using $\tiny{\beta=1}$. Furthermore, we leverage CLIP's image encoder to extract image embedding from the corresponding image $\mathbf{x}_{m-1}^i$ associated with each annotation $\mathbf{y}_{m-1}^i$. The image embedding is replicated $N$ times, corresponding to the number of objects in $\mathbf{y}_{m-1}^i$. The image embeddings are then concatenated with the grounding embeddings across the feature dimension. This process, aligned with the text prompt T, is employed in the generation process. It closely reproduces the realistic style and quality of the original images from the previous dataset $\mathcal{D}_{m-1}$. The image qualities that are varied with additional inputs are shown in Figure~\ref{fig:fig3_groundinginput}.

\subsection{Iterative class-wise refiner}
\label{subsec:refiner}
Despite our efforts to closely mimic the characteristics of real images, training the model with images $\mathcal{X}_{gen}$ generated using all of the previous annotations $\mathcal{Y}_{m-1}$ resulted in only limited performance improvements, while also leading to extended generation times. To address these issues, we employ a class-wise generation limit, denoting the maximum number of generated images for each class as $\mathcal{N}$. This constraint not only ensures a more efficient, but also a balanced generation process. 
We further employ a refinement process using the model $\mathcal{M}_{m-1}$ to ensure the quality and fidelity of the generated images. This refinement process is conducted through pseudo-labeling (described in \ref{subsec:fake_query}) where images containing objects with a probability lower than $p_\text{refine}$ are discarded in $\mathcal{D}_{gen}$. This approach presents a trade-off: a higher threshold $p_\text{refine}$ results in higher-quality images but often fails to meet the class-wise quantity $\mathcal{N}$, while a lower threshold achieves the quantity goal but compromises image fidelity. Therefore, we adopt an iterative process where the threshold $p_\text{refine}$ is gradually decreased by 0.05 in each generation cycle. This process continues until the generated images for all classes meet the pre-defined class-wise quantity $\mathcal{N}$ or until the lower-bound threshold $p_\text{refine}$, which is 0.4 in our case, is reached. 

\begin{figure}[t]
\centering{
\includegraphics[width=1.\linewidth]{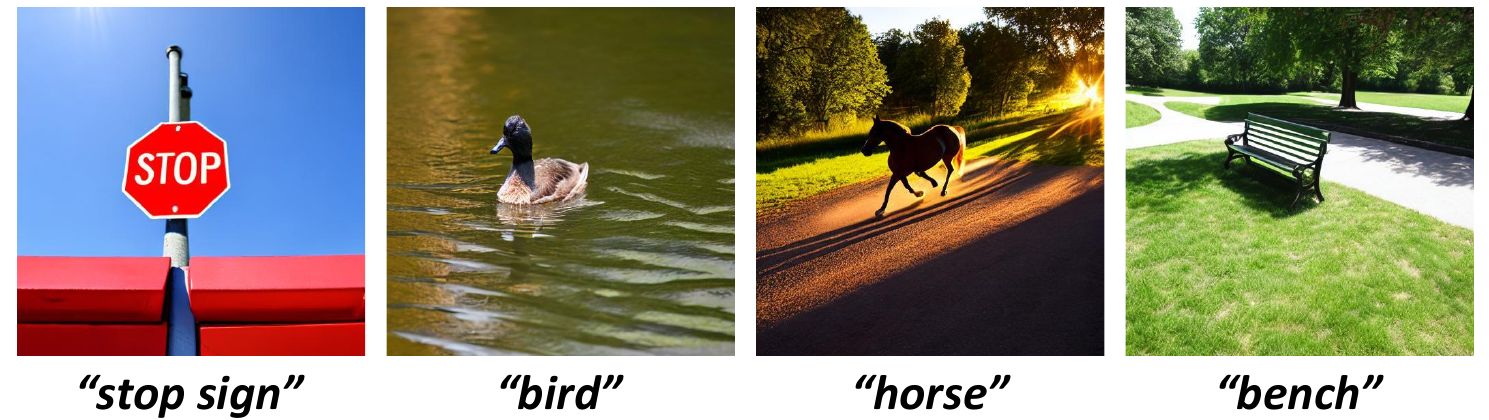}
}
    \vspace{-1em}
    \tiny\caption{Class-specific image generation. This process utilizes a single class label as a prompt and grounding input with a fixed location. For example, in the first column, we have $\mathbf{T}$ = ``stop sign", $\{\mathbf{C}_\text{label}$, $\mathbf{B}_\text{location}\}$ = \{(``stop sign", [0.3, 0.3, 0.6, 0.6])\}.}
\label{fig:class-specific}
\end{figure}

However, if after the completion of the iterative refinement process, the generated image count for any class still does not meet the standard $\mathcal{N}$, we introduce a class-specific generation strategy. This additional step involves creating an additional generation process with specialized prompts and phrases tailored for the classes that have not generated enough images. To focus the synthesis on the target object, we strategically position the bounding box at the center of normalized coordinates [0.3, 0.3, 0.6, 0.6]. In this manner, we also apply the previously described refinement process, utilizing $\mathcal{M}_{m-1}$ with only lower-bound threshold. Figure~\ref{fig:class-specific} shows an example of this class-specific image generation technique.

\subsection{Pseudo labeling}
\label{subsec:fake_query}
Based on the prediction mechanism introduced in~\cite{liu2023CLDETR, gupta2022ow}, D-DETR utilizes learned object queries in the decoder. Each query is processed through multiple decoder layers and then fed into the classification and regression heads to predict classes and bounding boxes, respectively. The classification head computes a matrix  $F_{\text{cls}} \in \mathbb{R}^{Q \times \text{(cls+1)}}$, where $Q$ denotes the number of object queries and $\text{cls+1}$ represents the total number of learned classes with an additional one for the background prediction. Each entry in $F_{\text{cls}}$ denotes a logit score, signifying the probability of a query being a specific class. 
For pseudo-labeling, after processing through the final decoder layer, we examine the logits in $F_{\text{cls}}$ for each query, identifying the logit with the highest score. If this highest score exceeds the pre-defined threshold $p_\text{pseudo}$, the query is then pseudo-labeled with the corresponding class. Concurrently, the regression head outputs a matrix $F_{\text{reg}} \in \mathbb{R}^{Q \times 4}$, where each column contains the normalized coordinates of the bbox for each query. The queries exceeding the threshold in $F_{\text{cls}}$ are aligned with the bbox predictions in $F_{\text{reg}}$ for the pseudo ground truth. It plays a crucial role in mitigating the forgetting of previously learned objects during the current training phase $\mathcal{T}_{m}$, particularly by reducing the misclassification of these objects as background, especially in scenarios where previous annotations $\mathcal{Y}_{n-1}$ are not available. This strategy is also used in Section \ref{subsec:refiner} to refine the synthetic images.

\subsection{Training with generated image}
\label{subsec:training_process}
While employing synthetic images $\mathcal{X}_{gen}$, we observed that despite our attempts to preserve the previous knowledge, the performance improvement when training directly is insufficient compared to the state-of-the-art. 
To improve this, instead of using the synthetic images $\mathcal{X}_{gen}$ as direct inputs for training, we enforce the new model $\mathcal{M}_{m}$ to acquire previous knowledge indirectly from the previous model $\mathcal{M}_{m-1}$. Inspired by~\cite{feng2022erders}, we apply an L2 distillation loss to both the classification and regression outputs. The formulation of the distillation function in terms of the object queries $Q$ and at a given task index $m$ is defined as follows:
\begin{equation}
\small
    \mathcal{L}_{\text{cls}} = \frac{1}{Q \times C} \sum_{i=1}^{Q} \sum_{j=1}^{C} \left( F_{\text{cls}, m}^{ij} - F_{\text{cls}, (m-1)}^{ij} \right)^2,
\end{equation}
\begin{equation}
\small
    \mathcal{L}_{\text{reg}} = \frac{1}{Q \times 4} \sum_{i=1}^{Q} \sum_{k=1}^{4} \left( F_{\text{reg}, m}^{ik} - F_{\text{reg}, (m-1)}^{ik} \right)^2,
\end{equation}
where $F_{\text{cls}, m}^{ij}$ and $F_{\text{reg}, m}^{ik}$ represent the predicted scores for the $i$-th query's $j$-th class index and $k$-th bbox coordinate by the new model $\mathcal{M}_{m}$, and $F_{\text{cls}, (m-1)}^{ij}$ and $F_{\text{reg}, (m-1)}^{ik}$ are those predicted by the old model $\mathcal{M}_{m-1}$. This approach retains the predictive consistency of $\mathcal{M}_{m-1}$, thereby mitigating the forgetting. Furthermore, since the decoder in D-DETR extracts predictions over 6 layers, we extend the application of L2 distillation loss across all these layers to facilitate distillation. In the training, we use the same loss formulation for D-DETR, denoted as $\mathcal{L}_{DETR}$. To effectively integrate and balance the distillation loss with the inherent D-DETR loss, we introduce a weight $\lambda$. The final loss function, reflecting a blend of the standard losses with the additional distillation components, is formalized as follows:
\begin{equation}
\small
    \mathcal{L}_{total} = \mathcal{L}_{DETR} + \lambda ( \alpha \mathcal{L}_{\text{cls}} + \beta \mathcal{L}_{\text{reg}} ).
\end{equation}\textbf{}
Here, $\alpha$ and $\beta$  are the weights for the classification and regression loss terms, respectively, adapted from the original D-DETR configuration, where $\alpha$ is 2 and $\beta$ is 5.
\begin{table*}[h]
\centering
\caption{CIOD results (\%) on COCO 2017 in \emph{two-phase setting}. The results of related research \cite{li2017learning, li2019rilod, peng2021sid, feng2022erders} extract from CL-DETR paper. The order of data follows the~\cite{feng2022erders}. The best performance is highlighted in \textbf{bold}, and a red upward arrow \textcolor{red}{↑}  signifies an improvement in performance relative to the state-of-the-art.}
\renewcommand{\arraystretch}{1.0}
\small
\resizebox{.98\textwidth}{!}{
\begin{tabular}{c|l|c|cccccc}
\hline\hline
{Scenarios} & \multicolumn{1}{c|}{{Method}}  & {Baseline} & $AP$ & $AP_{.5}$ & $AP_{.75}$ & $AP_{S}$ & $AP_{M}$ & $AP_{L}$ \\ \hline
\multirow{6}{*}{40 + 40} 
     & LWF~\cite{li2017learning} & GFLv1 & 17.2 & 25.4 & 18.6 & 7.9 & 18.4 & 24.3 \\ 
     & RILOD~\cite{li2019rilod} & GFLv1 & 29.9 & 45.0 & 32.0 & 15.8 & 33.0 & 40.5 \\ 
     & SID~\cite{peng2021sid} & GFLv1 & 34.0 & 51.4 & 36.3 & 18.4 & 38.4 & 44.9 \\ 
     & ERD~\cite{feng2022erders} & GFLv1 & 36.9 & 54.5 & 39.6 & 21.3 & 40.4 & 47.5 \\ 
     & CL-DETR~\cite{liu2023CLDETR} & Deformable DETR & 42.0 & 60.1 & 45.9 & 24.0 & 45.3& 55.6 \\ 
     & Ours & Deformable DETR & \textbf{43.0} \small\textcolor{red}{↑1.0} & \textbf{62.1} \small\textcolor{red}{↑2.0} & \textbf{47.1} \small\textcolor{red}{↑1.2} & \textbf{24.9} \small\textcolor{red}{↑0.9} & \textbf{46.9} \small\textcolor{red}{↑1.6} & \textbf{57.0} \small\textcolor{red}{↑1.4} \\ \hline 
\multirow{6}{*}{70 + 10} 
     & LWF~\cite{li2017learning} & GFLv1 & 7.1 & 12.4 & 7.0 & 4.8 & 9.5 & 10.0 \\ 
     & RILOD~\cite{li2019rilod} & GFLv1 & 24.5 & 37.9 & 25.7 & 14.2 & 27.4 & 33.5 \\ 
     & SID~\cite{peng2021sid} & GFLv1 & 32.8 & 49.0 & 35.0 & 17.1 & 36.9 & 44.5 \\ 
     & ERD~\cite{feng2022erders} & GFLv1 & 34.9 & 51.9 & 37.4 & 18.7 & 38.8 & 45.5 \\ 
     & CL-DETR~\cite{liu2023CLDETR} & Deformable DETR & 40.4 & 58.0 & 43.9 & 23.8 & 43.6 & 53.5 \\  
     & Ours & Deformable DETR & \textbf{40.9} \textcolor{red}{\small{↑0.5}} & \textbf{59.5} \textcolor{red}{\small{↑1.5}} & \textbf{44.8} \textcolor{red}{\small{↑0.9}} & \textbf{23.9} \textcolor{red}{\small{↑0.1}} & \textbf{44.7} \textcolor{red}{\small{↑1.1}} & \textbf{54.0} \textcolor{red}{\small{↑0.5}}  \\  \hline \hline 
\end{tabular}}
\label{table:table-1(main)}
\end{table*}
\begin{table*}[ht]
\centering
\renewcommand{\arraystretch}{1.1}
\caption{CIOD results ($AP$/$AP_{.5}$, \%) on COCO 2017 in \emph{multi-phase setting}. The tasks are divided into two scenarios: 40+10+10+10+10 and 40+20+20. Ours and CL-DETR are based on deformable DETR, while ERD, RILOD, and SID are based on GFLv1. A red upward arrow \textcolor{red}{↑} indicates a performance improvement compared to the state-of-the-art CL-DETR. The "-" symbol indicates a missing value, as reported in paper~\cite{liu2023CLDETR}.}
\label{tab:multi_results}
\resizebox{2.0\columnwidth}{!}{%
\begin{tabular}{l|c|cccc|cc}
\toprule
\multirow{2}{*}{\textbf{Method}} & \multirow{2}{*}{$\mathcal{T}_1$ (1-40)} & \multicolumn{4}{c|}{40+10+10+10+10} & \multicolumn{2}{c}{40+20+20} \\
\cline{3-8}
&  & $\mathcal{T}_2$ (40-50) & $\mathcal{T}_3$ (50-60) & $\mathcal{T}_4$ (60-70) & $\mathcal{T}_5$ (70-80) & $\mathcal{T}_2$ (40-60) & $\mathcal{T}_3$ (60-80) \\
\midrule
    Ours & \multirow{2}{*}{46.5 / 68.6} &  42.3 / 62.8  & 40.6 / 60.2  & 40.0 / 59.0  & \textbf{36.8} \textcolor{red}{\small{↑8.7}} / 54.7 & 42.5 / 62.2 & \textbf{41.1}\textcolor{red}{\small{↑5.8}}  / 59.5 \\
    CL-DETR~\cite{liu2023CLDETR} &  & - & - & - & 28.1  / - & - & 35.3 / -     \\
\midrule
    ERD~\cite{feng2022erders} & \multirow{3}{*}{45.7 / 66.3} & 36.4 / 53.9 & 30.8 / 46.7 & 26.2 / 39.9  & 20.7 / 31.8 & 36.7 / 54.6 & 32.4 / 48.6 \\
    RILOD~\cite{li2019rilod} &  & 25.4 / 38.9  & 11.2 / 17.3  & 10.5 / 15.6  & 8.4 / 12.5 & 27.8 / 42.8 & 15.8 / 4.0 \\
    SID~\cite{peng2021sid} & & 34.6 / 52.1 & 24.1 / 38.0  & 14.6 / 23.0  & 12.6 / 23.3 & 34.0 / 51.8 & 23.8 / 36.5 \\
\bottomrule
\end{tabular}}
\end{table*}

%% file: sec/4_experiments.tex
\section{Experiments}
\label{exp}

\subsection{Dataset and metrics}
\label{sec:Dataseteval}
Our research utilizes the MS COCO 2017~\cite{lin2014microsoft}, which consists of 80 diverse classes across 118,000 images for training and 5,000 images for evaluation. These classes are strategically divided based on our experiment scenario. For evaluation, we employ standard COCO metrics, including mean average precision (mAP, \%) at different intersection over union (IoU) thresholds and object sizes: $AP$, $AP_{.5}$, $AP_{.75}$, $AP_{S}$, $AP_{M}$, and $AP_{L}$. Here, $AP$ refers to the mAP calculated over IOU thresholds ranging from 0.5 to 0.95. In our ablation study, we introduce the \emph{forgetting percentage points} (FPP) as proposed by CL-DETR~\cite{liu2023CLDETR}, as a metric to evaluate the degree of forgetting for trained categories.

\subsection{Implementation and experiments}
\noindent{\textbf{Implementation details.}} Our method is based on deformable-DETR~\cite{zhu2020deformable}, which leverages pre-trained ResNet-50~\cite{he2016deep} as a multi-scale backbone. We set the number of object queries $Q$ to 300 while keeping all other settings consistent with our baseline~\cite{zhu2020deformable}. All experiments are performed using NVIDIA A100 GPUs with a batch size of 8. In our generation process, we utilize stable diffusion version 1.4. For incorporating grounding input, we employ pre-trained GLIGEN's gated self-attention weights that have been trained on various datasets including GoldG~\cite{li2022GOLDG}, O365~\cite{shao2019objects365}, SBU~\cite{ordonez2011SBU}, and CC3M~\cite{sharma2018cc3m}. Additionally, we set the classifier-free guidance scale to 7.5. 

\noindent{\textbf{Scenario setup.}} 
In our experiment, we focus on two scenarios: the \emph{two-phase setting} and the \emph{multiple-phase setting}. In the \emph{two-phase setting}, we train a model first task on $\mathcal{T}_1$ and then on a different task $\mathcal{T}_2$, evaluating on a combined total of $\mathcal{T}_1+\mathcal{T}_2$ classes, such as 40+40 or 70+10. In the \emph{multiple-phase setting}, we begin by training on 40 classes as a $\mathcal{T}_1$, then sequentially add new classes in $\mathcal{T}_n$ phases (e.g., 40+20+20 or 40+10+10+10+10), with evaluation conducted on all classes $\mathcal{T}_{1:n}$ learned up to each phase.

\subsection{Results}
\noindent{\textbf{Two-phase setting.}} 
Tab.~\ref{table:table-1(main)} shows that our method outperforms previous approaches such as LWF~\cite{li2017learning}, RILOD~\cite{li2019rilod}, SID~\cite{peng2021sid}, and ERD~\cite{feng2022erders} using GFLv1~\cite{li2020generalized}, including CL-DETR~\cite{liu2023CLDETR}. Importantly, we achieved a 0.5\% increase in $AP$ for the 70+10 scenario and 1.0\% for the 40+40 scenario. Moreover, we observed even higher gains of 1.5\%  and 2.0\% in $AP_{.5}$, respectively. It is particularly noteworthy that while CL-DETR relies on a 10\% replay buffer comprising real data from previous tasks, our method stands out by achieving remarkable performance improvements without any reliance on real previous data.

\noindent{\textbf{Multi-phase setting.}}
Tab.~\ref{tab:multi_results} shows that our method, which utilizes synthetic image-based training, surprisingly outperforms other approaches significantly in multi-phase scenarios. Despite using different baselines like~\cite{feng2022erders, li2019rilod, peng2021sid}, it is evident that our method maintains consistent performance. We achieve 8.7\% and 5.8\% gains in $AP$ for the 40+10+10+10+10 and 40+20+20 scenarios, respectively, compared to CL-DETR. This highlights that our approach, which uniformly employs synthetic data through knowledge distillation from the old model across all phases, effectively trains on new task data while maintaining high performance by alleviating catastrophic forgetting.

\subsection{Ablations}
\noindent{\textbf{Main components.}}
In Tab.~\ref{tab:ablation_maina}, we present an ablation study of our method's components in the 70+10 scenario. For the `Fine-tuning' component, we do not apply specific CIOD strategies. The results show a significant improvement when employing the pseudo-labeling strategy, which notably reduces FPP by 39.1\% in $AP$. This highlights its crucial role in minimizing the misclassification of previously trained objects as background. Following the introduction of a synthetic dataset to mitigate forgetting, we noticed a modest increase in $AP$ by 1.2\% in all categories and a reduction in FPP by 1.5\%. Although this indicates that synthetic data contributes to knowledge retention, its impact on reaching state-of-the-art performance is somewhat insufficient. However, the results take a significant turn when we integrate distillation with the synthetic dataset training, marking a substantial improvement. As a result, we achieve an AP of 40.9\% in all categories and 41.5\% in old categories, along with an FPP of 1.9\%. These indicate a significant advancement in the effectiveness of our method.

\vspace{0.1em}
\noindent{\textbf{Pseudo-labeling.}}
Tab.~\ref{tab:ablation-pseudo labeling} illustrates the impact of varying confidence score $p_\text{pseudo}$ thresholds on the selection of optimal queries for pseudo ground-truth labeling. The data reveals a marked improvement in performance when predictions are labeled using a query score threshold above 0.3. However, it also shows a gradual decline in performance as the threshold is increased beyond this point.

\vspace{0.1em}
\noindent{\textbf{Refiner.}}
Tab.~\ref{tab:ablation_study_category_threshold} presents our findings on how varying the number of generated images per class ($\mathcal{N}$) and the refinement threshold ($p_\text{refine}$) influences the performance of our iterative refinement method (Section~\ref{subsec:refiner}).

\begin{table*}[t]
\centering
\small
\renewcommand{\arraystretch}{1.0}
\caption{Ablation study of main contribution components on COCO 2017 (\emph{two-phase setting}, 70+10). The metrics assess performance after completing training across all phases, measuring results across all categories (higher is better) and specifically in old categories (higher is better). The \emph{forgetting percentage point} (FPP, lower is better) specifically reflects the performance change in the initial 70 categories, as measured by the difference in $AP$ between the first phase and the last phase. The best performance is represented in \textbf{bold}, with the final row indicating our method's results.}
\label{tab:ablation_maina}
\resizebox{.9\textwidth}{!}{%
\begin{tabular}{lccccccccc}

\toprule
\multicolumn{1}{c}{\multirow{2}{*}{Method}} & \multicolumn{3}{c}{All categories \( \uparrow \)} & \multicolumn{3}{c}{Old categories \( \uparrow \)} & \multicolumn{3}{c}{FPP \( \downarrow \)} \\
\cmidrule(lr){2-4} \cmidrule(lr){5-7} \cmidrule(lr){8-10}
     & $AP$ & $AP_{.5}$ & $AP_{.75}$  & $AP$ & $AP_{.5}$ & $AP_{.75}$ & $AP$ & $AP_{.5}$ & $AP_{.75}$\\
\midrule
    Fine-tuning      & 14.8 & 23.6 & 15.6 & 0.0 & 0.0 & 0.0 & 43.4 & 62.8 & 47.2 \\
    + Pseudo labeling      & 38.6 & 56.2 & 42.1 & 39.1 & 57.3 & 42.7 & 4.3 & 5.5 & 4.5 \\
    ++ Deep generative replay   & 39.8 & 57.7 & 43.4 & 40.6 & 59.2 & 44.0 & 2.8 & 3.6 & 3.2 \\ \rowcolor{lightgray}
    +++ Knowledge distillation  & \textbf{40.9} & \textbf{59.5} & \textbf{44.8} & \textbf{41.5} & \textbf{60.6} & \textbf{45.4} & \textbf{1.9} & \textbf{2.2} & \textbf{1.8}\\
\bottomrule
\end{tabular}%
}
\end{table*}
\begin{table}[ht]
\centering
\small
\caption{Ablation study of the range of confidence scores in the pseudo-labeling strategy on COCO 2017 (70+10). The best performance is highlighted in \textbf{bold}.}
\label{tab:ablation-pseudo labeling}
\renewcommand{\arraystretch}{1.0}
\resizebox{0.47\textwidth}{!}{%
\begin{tabular}{ccccccc}
    \toprule
    Setting & $AP$ & $AP_{.5}$ & $AP_{.75}$ & $AP_{S}$ & $AP_{M}$ & $AP_{L}$ \\
    \midrule
    $p_\text{pseudo} \geq 0.2$ & 29.5 & 44.9 & 32.0 & 17.4 & 33.4 & 38.8 \\ \rowcolor{lightgray}
    $p_\text{pseudo} \geq 0.3$ & \textbf{38.6} & \textbf{56.2} & \textbf{42.1} & \textbf{22.3} & \textbf{42.1} & \textbf{50.6} \\
    $p_\text{pseudo} \geq 0.4$ & 37.5 & 54.2 & 40.8 & 22.0 & 41.3 & 48.5 \\
    $p_\text{pseudo} \geq 0.5$ & 35.2 & 51.1 & 38.8 & 20.3 & 38.7 & 46.1 \\
    \bottomrule
\end{tabular}}
\end{table}
\begin{table}[ht]
\centering
\small
\caption{Ablation study on image generation regulation and refinement confidence score thresholds on COCO 2017 (70+10). The best result is highlighted in \textbf{bold} among each ablation.}
\label{tab:ablation_study_category_threshold}
\renewcommand{\arraystretch}{1.0}
\resizebox{0.47\textwidth}{!}{%
\begin{tabular}{ccccccc}
\toprule
    Setting & $AP$ & $AP_{.5}$ & $AP_{.75}$ & $AP_{S}$ & $AP_{M}$ & $AP_{L}$ \\ \rowcolor{lightgray}
\midrule
    $\mathcal{N}= 50$      & \textbf{40.9} & \textbf{59.5} & \textbf{44.8} & \textbf{24.0} & \textbf{44.7} & 54.0 \\
    $\mathcal{N}=100$     & 40.7 & 59.4 & 44.6 & \textbf{24.0} & 44.4 & 53.9 \\
    $\mathcal{N}=200$     & 40.8 & \textbf{59.5} & 44.6 & \textbf{24.0} & 44.2 & \textbf{54.5} \\ 
    $\mathcal{N}= \infty$  & 39.9 & 58.6 & 43.4 & 23.0 & 43.6 & 53.1 \\ \hline
    $p_\text{refine} = 0.4$ & 39.8 & 58.8 & 43.4 & 23.1 & 43.2 & 52.6 \\
    $p_\text{refine} = 0.8$ & 38.5 & 57.2 & 42.2 & 22.9 & 41.8 & 51.0 \\\rowcolor{lightgray}
    $p_\text{refine} \in [0.4,0.8]$ & \textbf{40.9} & \textbf{59.5} & \textbf{44.8} & 23.9 & \textbf{44.7} & \textbf{54.0} \\
    $p_\text{refine} \in [0.5,0.8]$ & 40.7 & 59.3 & 44.5 & \textbf{24.1} & 44.0 & 53.6 \\ 
    $p_\text{refine} \in [0.6,0.8]$ & 40.5 & 59.2 & 44.3 & 23.4 & 44.2 & 53.6 \\
    $p_\text{refine} \in [0.7,0.8]$ & 39.6 & 56.8 & 43.3 & 22.5 & 42.6 & 51.8 \\
\bottomrule
\end{tabular}}
\end{table}
When examining different $\mathcal{N}$ values (50, 100, and 200), we observed comparable performances, particularly with a fixed threshold range (e.g., from 0.8 to 0.4). This suggests that our method is robust to variations in class-wise image count regulation. However, when we set $\mathcal{N}$ to "no limit" ($\infty$), generating images based on all old annotations without class-wise restrictions, there is a performance drop of 1\% (39.9\%) in $AP$ compared to our best (40.9\%). This outcome highlights the necessity of regulating image generation for each class. Furthermore, by limiting the production quantity per class using $\mathcal{N}$, we significantly reduce the time required for generation. This efficiency gain is further discussed in the supplementary Tab. 2.

Regarding the refinement threshold ($p_\text{refine}$), our best result (40.9\%) is obtained with a dynamic range between 0.4 and 0.8. Setting $p_\text{refine}$ to a fixed value, either at the low end (0.4) or high end (0.8), led to diminished performance. This indicates the importance of a dynamic threshold range in optimizing the generation and refinement process. 
In conclusion, these results demonstrate that our iterative refinement strategy effectively refines the synthetic images while balancing the quality and quantity of the synthetic images.

\begin{table}[t]
\centering
\small
\caption{Ablation study of knowledge distillation weight on COCO 2017 (70+10). The best result is highlighted in \textbf{bold}.}
\label{tab:ablation-KDweight}
\begin{tabular}{ccccccc}
    \toprule
    Weight & $AP$ & $AP_{.5}$ & $AP_{.75}$ & $AP_{S}$ & $AP_{M}$ & $AP_{L}$ \\
    \midrule
        $\lambda=1$ & 40.6 & 59.2 & 44.2 & 23. 7& 44.1 & 53.8 \\
        \rowcolor{lightgray}$\lambda=2$ & \textbf{40.9 }& \textbf{59.5} & \textbf{44.8} & \textbf{23.9} & \textbf{44.7} & \textbf{54.0} \\
        $\lambda=3$ & 40.5 & 59.3 & 44.4 & 24.0 & 44.1 & 53.0 \\
        \bottomrule
\end{tabular}
\vspace{-0.5em}
\end{table}
\vspace{0.1em}
\noindent{\textbf{Weight parameter $\lambda$.}}
In Tab.~\ref{tab:ablation-KDweight}, we examine the effect of different weight parameter $\lambda$. We found that a weight of 2 achieved the best performance with an AP of 40.9\%. Importantly, all tested weights performed better than the current state-of-the-art performance of 40.4\%, demonstrating the effectiveness of our knowledge distillation approach using synthetic images.

\vspace{0.1em}
\noindent\textbf{CLIP image embedding.}
In Tab~\ref{tab:sub_CLIPImageEmbedding}, we conducted an ablation experiment to evaluate the effect of incorporating CLIP's image embedding in the generation process (Sec.~\ref{subsec:generate process}). The result indicates that incorporating CLIP's image embedding led to a performance improvement of 1.2\% in $AP$. This highlights the impact of CLIP's image embedding in enhancing the realism of synthetic images, which in turn positively impacts the detector performance. On the other hand, without the CLIP's image embedding, it results in an inferior performance that is similar to the baseline using pseudo-labeling alone (38.6\% $AP$ in Tab.~\ref{tab:ablation_maina}). This result implies that image quality and realism are important in CIOD with the deep generative model to effectively prevent the catastrophic forgetting.
\begin{table}[t!]
\centering
\small
\caption{Ablation study of CLIP's image embedding on COCO 2017 (70+10). The experiment was conducted excluding L2 knowledge distillation to evaluate the effect of synthetic data. The best result is highlighted in \textbf{bold}. A red upward arrow \textcolor{red}{↑} indicates the performance improvement.}

\label{tab:sub_CLIPImageEmbedding}
\begin{tabular}{ccccccc}
    \toprule
        Setting & $AP$ & $AP_{.5}$ & $AP_{.75}$\\
    \midrule
        w/o Image embedding & 38.6 & 56.9 & 42.1  \\
        \rowcolor{lightgray}w/ Image embedding & \textbf{39.8}~\textcolor{red}{1.2↑} & \textbf{57.7} \textcolor{red}{0.8↑} & \textbf{43.4} \textcolor{red}{1.3↑} \\
        \bottomrule
\end{tabular}
\end{table}

%% file: sec/5_results.tex
\section{Conclusions}
In this paper, we introduced the SDDGR strategy, a novel diffusion-based deep generative replay approach for class incremental object detection. The proposed SDDGR includes a method for generating synthetic images that encompass objects from previously trained classes, with the goal of enhancing their quality and high fidelity while maintaining computational efficiency. To achieve this, we suggested a rigorous refinement technique and class-wise regulation of quantity. Additionally, the synthetic images are effectively used to mitigate forgetting through the application of L2 knowledge distillation. Finally, SDDGR utilizes an effective pseudo-labeling technique that substantially reduces the misclassification of objects as background. The combination of these proposed methods enables our SDDGR to achieve state-of-the-art performance in class incremental object detection. 
